%% file: main.tex
\def\year{2022}\relax
\documentclass[letterpaper]{article} 
\usepackage{aaai22}  
\usepackage{times}  
\usepackage{helvet}  
\usepackage{courier}  
\usepackage[hyphens]{url}  
\usepackage{graphicx} 
\usepackage{natbib}  
\usepackage{caption} 
\DeclareCaptionStyle{ruled}{labelfont=normalfont,labelsep=colon,strut=off} 
\usepackage{algorithm}
\usepackage{algorithmic}

\usepackage{newfloat}
\usepackage{listings}
\floatstyle{ruled}
\newfloat{listing}{tb}{lst}{}
\floatname{listing}{Listing}

\usepackage[hybrid]{markdown}
\usepackage{amsmath}
\usepackage{amssymb}
\usepackage{graphicx}
\usepackage{amsmath}
\usepackage{amssymb}
\usepackage{amsthm}
\usepackage{colortbl}
\usepackage{multirow}
\usepackage{stmaryrd}
\theoremstyle{definition}
\usepackage{xspace}

\usepackage{booktabs}
\usepackage{siunitx}
\usepackage{tipa}
\usepackage{pifont}
\usepackage{amsbsy}
\usepackage{colortbl}

\newtheorem{proposition}{Proposition}
\usepackage{mathtools}
\usepackage{nicematrix}

\newcommand{\modelname}{\texttt{FuzzQE}\xspace}
\newcommand{\luka}{Łukasiewicz}
\newcommand{\godel}{Gödel}

\newcommand{\be}{{\mathbf{p}_e}}

\newcommand{\bsq}{{\mathbf{S}_q}}
\newcommand{\sq}{{S_q}}
\newcommand{\tnorm}{$t$-norm}
\newcommand{\tnorms}{$t$-norms}
\newcommand{\tconorm}{$t$-conorm}
\newcommand{\tconorms}{$t$-conorms}

\newcommand{\conjop}{\mathcal{C}}
\newcommand{\disjop}{\mathcal{D}}
\newcommand{\negop}{\mathcal{N}}
\newcommand{\myroman}[1]{\uppercase\expandafter{\romannumeral#1}}

\newcommand\clearrow{\global\let\rowmac\relax}

\newcommand{\xmark}{\ding{55}}  

\newcommand{\nop}[1]{}

\title{Fuzzy Logic Based Logical Query Answering on Knowledge Graphs}
\author{
    Xuelu Chen,
    Ziniu Hu,
    Yizhou Sun
}
\affiliations{
    Department of Computer Science, University of California, Los Angeles\\
    \{shirleychen, bull, yzsun\}@ucla.edu\\
}

\begin{document}

\maketitle

\input{writeup/0-abstract}
\input{writeup/1-introduction}

\input{writeup/2-relatedwork}
\input{writeup/4-methodology}

\input{writeup/5-experiments}

\input{writeup/6-conclusion}

\section*{Acknowledgements}
This work was partially supported by NSF III-1705169, NSF 1937599, DARPA HR00112090027, Okawa Foundation Grant, Amazon Research Awards, Cisco research grant, and Picsart gift.
\bibliography{ref}
\input{writeup/7-0-appendix}

\end{document}

%% file: writeup/0-abstract.tex
\begin{abstract}
Answering complex First-Order Logical (FOL) queries on large-scale incomplete knowledge graphs (KGs) is an important yet challenging task.
Recent advances embed logical queries and KG entities in the same space and conduct query answering via dense similarity search.
However, most logical operators designed in previous studies do not satisfy the axiomatic system of classical logic, limiting their performance. Moreover, these logical operators are parameterized and thus require many complex FOL queries as training data, which are often arduous to collect or even inaccessible in most real-world KGs.
We thus present \modelname
, a fuzzy logic based logical query embedding framework for answering FOL queries over KGs. 
\modelname follows fuzzy logic to define logical operators in a principled and learning-free manner, where only entity and relation embeddings require learning.
FuzzQE can further benefit from labeled complex logical queries for training.
Extensive experiments on two benchmark datasets demonstrate that \modelname~provides significantly better performance
in answering FOL queries compared to state-of-the-art methods. In addition, \modelname trained with only KG link prediction can achieve comparable performance to those trained with extra complex query data.

\end{abstract}

%% file: writeup/1-introduction.tex
\section{Introduction}







\begin{figure}[t]
    \centering
    \includegraphics[width=\linewidth]{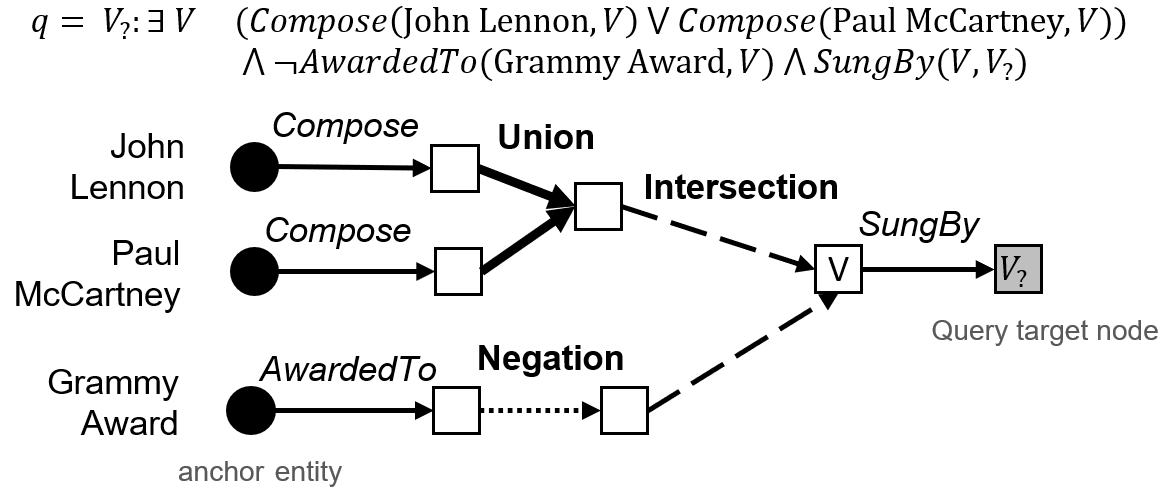}
    \caption{FOL query and its dependency graph for the question ``Who sang the songs that were written by John Lennon or Paul McCartney but never won a Grammy Award?''.} 
    \label{fig:computation_graph}
\end{figure}

Knowledge graphs (KGs), such as Freebase \citep{bollacker2008freebase}, YAGO \citep{rebele2016yago}, and NELL \citep{mitchell2018never}, provide structured representations of facts about real-world entities and relations. One of the fundamental tasks over KGs is to answer complex queries involving logical reasoning, e.g., 
answering First-Order Logical (FOL) queries with existential quantification ($\exists$), conjunction ($\land$), disjunction ($\lor$), and negation ($\neg$).
For instance, the question ``Who sang the songs that were written by John Lennon or Paul McCartney but never won a Grammy Award?'' can be expressed as the FOL query shown in Fig \ref{fig:computation_graph}.

This task is challenging due to the size and the incompleteness of KGs.
FOL query answering has been studied as a graph query optimization problem in the database community \cite{hartig2007sparql, lei2011rdfsubgraph, schmidt2010foundations}.
These methods traverse the KG to retrieve answers for each sub-query and then merge the results. Though being extensively studied, these methods cannot well resolve the above-mentioned challenges.
The time complexity exponentially grows with the query complexity and is affected by the size of the intermediate results. This makes them difficult to scale to modern KGs, whose entities are often numbered in millions \cite{bollacker2008freebase, Wikidata2014}.
For example, 
Wikidata is one of the most influential KGs and reports that their query engine fails when the number of entities in a sub-query (e.g., people born in Germany) exceeds a certain threshold\footnote{\url{https://www.wikidata.org/wiki/Wikidata:SPARQL_query_service/query_optimization}}. 
In addition, real-world KGs are often incomplete,
which prevents directly answering many queries 
by searching KGs.
A recent study shows that only 0.5\% of football players in Wikidata have a highly complete profile,
while over 40\% contain only basic information
\cite{wikidatacompleteness2018}.

To address the challenges of time complexity and KG incompleteness, 
a line of recent studies \cite{GQE, Query2Box, BetaE} embed logical queries and entities into the same vector space.
The idea is to represent a query using a \emph{dependency graph} (Figure \ref{fig:computation_graph}) and embed a complex logical query by iteratively computing embeddings from \emph{anchor entities} to the \emph{target node} in a bottom-up manner. 
The continuous and meaningful entity embeddings empower these approaches to handle missing edges. In addition, these models significantly reduce time and space complexity for inference, as they reduce query answering to dense similarity matching of query and entity embeddings and can speed it up using methods like maximum inner product search (MIPS) \cite{MIPS}.

These methods nonetheless entail several limitations:
First, the logical operators in these models are often defined ad-hoc, and many do not satisfy basic logic laws 
(e.g., the associative law $ (\psi_1 \land \psi_2) \land \psi_3 \equiv \psi_1 \land (\psi_2 \land \psi_3)$ for logical formulae $\psi_1, \psi_2, \psi_3$), 
which limits their inference accuracy. 
Second, the logical operators of existing methods are based on deep architectures, which require many training queries containing such logical operations to learn the parameters. This greatly limits the models' scope of application due to the challenge of collecting numerous reasonable complex queries with accurate answers.

Our goal is to create a logical query embedding framework that satisfies logical laws and provides learning-free logical operators. 
We hereby present \modelname~(\underline{Fuzz}y \underline{Q}uery \underline{E}mbedding), a fuzzy logic based embedding framework for answering logical queries on KGs.
We borrow the idea of fuzzy logic and use fuzzy conjunction, disjunction, and negation 
to implement logical operators in a more principled and learning-free manner.
Our approach provides the following advantages over existing approaches:
(i)
\modelname~employs differentiable logical operators that fully satisfy the axioms of logical operations and can preserve logical operation properties in vector space. 
This superiority is corroborated by extensive experiments on two benchmark datasets, which demonstrate that \modelname~delivers a significantly better performance compared to state-of-the-art methods in answering FOL queries.
(ii)
Our logical operations do not require learning any operator-specific parameters. We conduct experiments to show that even when our model is only trained with link prediction, it achieves better results than state-of-the-art logical query embedding models trained with extra complex query data.
This represents a huge advantage in real-world applications since complex FOL training queries are often arduous to collect.
In addition, when complex training queries are available, the performance of \modelname can be further enhanced.

In addition to proposing this novel and effective framework, 
we propose some basic properties that a logical query embedding model ought to possess as well as analyze whether existing models can fulfill these conditions. 
This analysis provides theoretical guidance for future research on embedding-based logical query answering models.

%% file: writeup/2-relatedwork.tex

\section{Related Work}
Embedding entities in Knowledge Graphs (KGs) into continuous embeddings have been extensively studied~\cite{TransE, DistMult, ComplEx, RotatE}, which can answer one-hop relational queries via link prediction. These models, however, cannot handle queries with multi-hop~\cite{DBLP:conf/emnlp/GuML15} or complex logical reasoning. 
\citet{GQE} thus propose a graph-query embedding (GQE) framework that encodes a conjunctive query via a dependency graph with relation projection and conjunction ($\land$) as operators. 
\citet{Query2Box} extend GQE by using box embedding to represent entity sets, where they define the disjunction ($\lor$) operator to support Existential Positive First-Order (EPFO) queries.  \citet{DBLP:conf/nips/SunAB0C20} concurrently propose to represent sets as count-min sketch~\cite{DBLP:journals/jal/CormodeM05} that can support conjunction and disjunction operators. More recently, ~\citet{BetaE} further include the negation operator ($\neg$) by modeling the query and entity set as beta distributions. 
\citet{friedman2020UAI} extend FOL query answering to probabilistic databases.
These query embedding models have shown promising results to conduct multi-hop logical reasoning over incomplete KGs efficiently regarding time and space;
however, we found that these models do not satisfy the axioms of either Boolean logic \cite{chvalovsky2012independence} or fuzzy logic \cite{KlementTNormBook},
which limits their inference accuracy. 
To address this issue, our approach draws from fuzzy logic and uses the fuzzy conjunction, disjunction, and negation operations to define the logical operators in the vector space.

In addition to the above logical query embedding models, a recent work CQD \cite{CQD} proposes training an embedding-based KG completion model (e.g., ComplEx \cite{ComplEx}) to impute missing edges during inference and merge entity rankings using \emph{t-norms} and \emph{t-conorms} \cite{KlementTNormBook}.
Using beam search for inference, CQD has demonstrated strong capability of generalizing from KG edges to arbitrary EPFO 
queries.
However, 
CQD has severe scalability issues
since it involves scoring every entity for every atomic query. 
This is undesirable in real-world applications, since the number of entities in real-world KGs are often in millions \cite{bollacker2008freebase, Wikidata2014}. 
Furthermore, 
its inference accuracy is thus bounded by KG link prediction performance. 
In contrast, our model is highly scalable, and its performance can be further enhanced when additional complex queries are available for training.

%% file: writeup/4-methodology.tex
\input{writeup/3-preliminaries}

\section{Methodology}
In this section, we propose our model \modelname, a framework for answering FOL queries in the presence of missing edges.
\modelname~embeds 
queries as \emph{fuzzy vectors} \cite{FuzzyVector1977}. 
Logical operators are implemented via fuzzy conjunction, fuzzy disjunction and fuzzy negation in the embedding space.

\subsection{Queries and Entities in Fuzzy Space}
Predicting whether an entity can answer a query means predicting the probability that the entity belongs to the answer set of this query. 
In our work, we embed queries and entities to the \emph{fuzzy space}  $[0,1]^d$, a subspace of $\mathbb{R}^d$ \cite{FuzzyVector1977}.

\paragraph{Query Embedding} Consider a query $q$ and its fuzzy answer set $S_q$, its embedding $\bsq$ is defined as a fuzzy vector $\bsq\in [0,1]^d$ \cite{FuzzyVector1977}.
Intuitively, let $\Omega$ denote the universe of all the elements, and let $\{U_i\}_{i=1}^{d}$ denote a partition over $\Omega$, i.e., $\Omega = \cup_{i=1}^d{U_i}$ and $U_i\cap U_j = \emptyset$ for $i\neq j$. Each dimension $i$ of $\bsq$ denotes the probability whether the corresponding subset $U_i$ is part of the answer set $S_q$, i.e., $\bsq(i)=\Pr(U_i \subseteq \sq)$. 

\nop{
\begin{align*}
    & \bsq \in [0,1]^d, \quad \bsq(i)=\Pr(U_i \subseteq \sq) \\
    & \be \in [0,1]^d, \quad \be(i) = \Pr(e \in U_i), \sum_{i=1}^{d} \be(i)=1
\end{align*}
}

\paragraph{Entity Embedding} For an entity $e$, we consider its embedding $\be$ from the same fuzzy space, i.e., $\be \in [0,1]^d$. To model its uncertainty, we model it as a categorical distribution to fall into each subset $U_i$, namely, $\be(i) = \Pr(e \in U_i)$, and $\sum_{i=1}^{d} \be(i)=1$.
\paragraph{Score Function} Accordingly, the score function  $\phi(q,e)$ is defined as the expected probability that $e$ belongs to the fuzzy set $S_q$:
\begin{small}
\begin{align*}
    \phi(q,e) 
    & = \mathbb{E}_{e \sim \be}[e \in \sq] \\
    & = \sum_{i=1}^{d} \Pr(e \in U_i)  \Pr(U_i \subseteq \sq) \\
    & = \bsq^\intercal \be
\end{align*}
 \end{small}Note for query embedding in \modelname, the all-one vector $\mathbf{1}$ represents the universe set (i.e., $\Omega$), and the all-zero vector $\mathbf{0}$ represents an empty set $\emptyset$.

The above representation and scoring provides the following benefits:
(i) The representation is endowed with probabilistic interpretation, and
(ii) each dimension of the embedding vector is between $[0,1]$, which satisfies the domain and range requirements of fuzzy logic and allows the model to execute element-wise fuzzy conjunction/disjunction/negation.
In light of the fact that $L_1$ normalization could impose sparsity on entity embeddings, we also alternatively explore adopting $L_2$ normalization to improve embedding learning, i.e., $\sum_{i=1}^{d} {\mathbf{p}_e^2}(i)=1$.



\subsection{Relation Projection for Atomic Queries}
Atomic queries like $q=\emph{Compose}(\emph{John Lennon}, V_?)$ serve as building blocks of complex queries. To embed atomic queries, we associate each relation $r \in \mathcal{R}$ with a projection operator $\mathcal{P}_r$, which is modeled by a neural network with a weight matrix $\mathbf{W}_r \in \mathbb{R}^{d \times d}$ and a bias vector $\mathbf{b}_r \in \mathbb{R}^d$, and transforms an anchor entity embedding $\be$ into a query embedding:
\begin{equation*}
    \bsq = \mathcal{P}_r(\be) = \mathbf{g}( \text{LN}( \mathbf{W}_r \be + \mathbf{b}_r))
\end{equation*}
where $\text{LN}$ is Layer Normalization \cite{LayerNorm}, and
$\mathbf{g}: \mathbb{R}^d \mapsto [0,1]^d$ is a mapping function that constrains $\bsq \in [0,1]^d$. 
Particularly, we consider two different choices for $\mathbf{g}$:
\begin{align*}
    \text{Logistic function}:
    &~\mathbf{g}(\mathbf{x}) = \frac{1}{1+ e ^{-(\mathbf{x})}} \\
    \text{Bounded rectifier}:
    &~\mathbf{g}(\mathbf{x}) = \min(\max(\mathbf{x},0), 1)
\end{align*}
We follow \cite{RGCN} and adopt \emph{basis-decomposition} to define $\mathbf{W}_r$ and $\mathbf{b}_r$: 
\begin{equation*}
    \mathbf{W}_r = \sum_{j=1}^{K} \alpha_{rj}\mathbf{M}_j;
    \quad \mathbf{b}_r = \sum_{j=1}^{K} \alpha_{rj}\mathbf{v}_j
\end{equation*}
Namely, $\mathbf{W}_r$ as a linear combination of $K$ basis transformations $\mathbf{M}_j \in \mathbb{R}^{d \times d}$ with coefficients $\alpha_{rj}$ that depend on $r$. Similarly, $\mathbf{b}_r$ is a linear combination of $K$ basis vectors $\mathbf{v}_j \in \mathbb{R}^d$ with coefficients $\alpha_{rj}$.
This form prevents the rapid growth in the number of parameters with the number of relations and alleviates overfitting on rare relations. It can be seen as a form of effective weight sharing among different relation types \cite{RGCN}. Atomic queries that project from one set to another can be embedded similarly.

In principle, any sufficiently expressive neural network or translation-based KG embedding model \cite{TransE, ji2015knowledge} could be employed as the relation projection operator in our framework.



\subsection{Fuzzy Logic Based Logical Operators} 
Fuzzy logic is mathematically equivalent to fuzzy set theory \cite{KlirFuzzyBook}, with fuzzy conjunction equivalent to fuzzy set intersection, fuzzy disjunction equivalent to fuzzy set union, and fuzzy negation to fuzzy set complement.
Fuzzy logic could thus be used to define operations over fuzzy vectors.
As discussed in Section \ref{sec:fuzzylogic}, the three most prominent \tnorm~based logic systems are product logic, \godel~logic, and \luka~logic \cite{KlementTNormBook}.
With reference to product logic, \modelname~computes the embeddings of $q_1 \land q_2$, $q_1 \lor q_2$, and $\neg q$ as follows:
\begin{alignat*}{2}
        q_1 \land q_2: & \quad \conjop(\bsq_1, \bsq_2) && =   \bsq_1 \circ \bsq_2 \\
        q_1 \lor q_2: & \quad \disjop(\bsq_1, \bsq_2) && = \bsq_1 + \bsq_2 - \bsq_1 \circ \bsq_2 \\
        \neg q: & \quad \negop(\bsq) && = \mathbf{1} - \bsq
\end{alignat*}
where 
$\circ$ denotes element-wise multiplication (fuzzy conjunction), $\mathbf{1}$ is the all-one vector, and $\conjop, \disjop, \negop$ denote the embedding based logical operators respectively. 

Alternatively, the conjunction and disjuction operators can be designed based on \godel~logic as follows:
\begin{alignat*}{2}
    q_1 \land q_2: & \quad \conjop(\bsq_1, \bsq_2) && =  \min( \bsq_1, \bsq_2) \\
    q_1 \lor q_2: & \quad \disjop(\bsq_1, \bsq_2) && = \max (\bsq_1, \bsq_2)
\end{alignat*}
where $\min, \max$ denotes element-wise minimum and maximum operation respectively. 

We omit \luka~logic here since its output domain is heavily concentrated in $\{0,1\}$, which causes a query embedding learning problem. 
More discussions about these three logic systems can be found in Appendix \ref{ap:tnorm}.



\begin{figure*}[h]
    \centering
    \includegraphics[width=\linewidth]{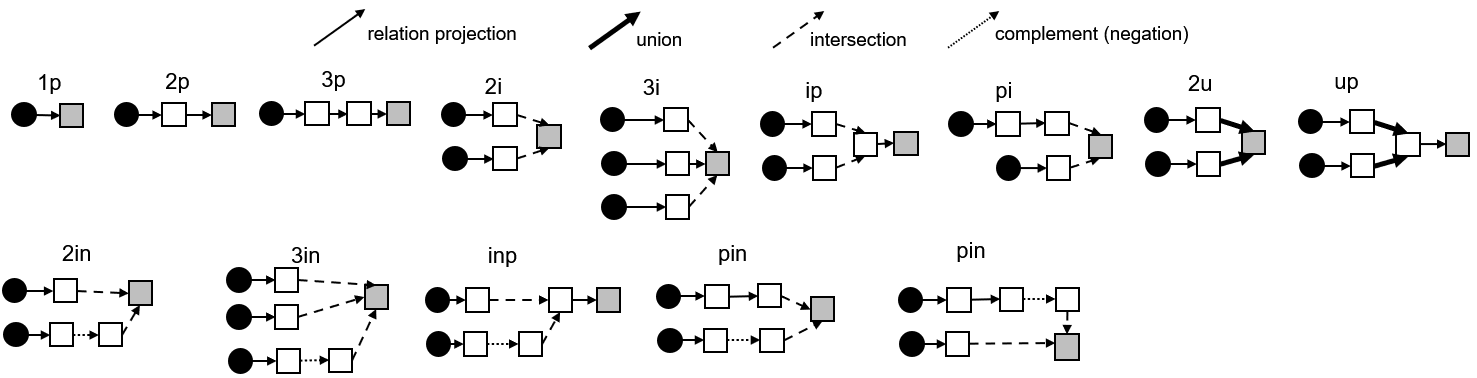}
    \caption{Query structure types in the datasets provided by BetaE \cite{BetaE}. Naming convention: $p$ for relation projection, $i$ for conjunction (intersection), $n$ for negation (complement), $u$ for disjunction (union). $10$ query structures are used for both training and evaluation: \emph{1p, 2p, 3p, 2i, 3i, 2in, 3in, inp, pni, pin}. In order to test the generalization capability of the model, 4 query structures (\emph{ip, pi, 2u, up}) are not present in training and only included for evaluation. } 
    \label{fig:query_structure}
\end{figure*}

\subsection{Model Learning and Inference}
Given a query $q$, we optimize the following objective:
\small
\begin{equation*}
    L= - \log \sigma(\frac{1}{Z_{q}}\phi(q,e) - \gamma) - \frac{1}{k} \sum_{i=1}^{k} \log \sigma(\gamma - \frac{1}{Z_{q}}\phi(q,e')) 
\end{equation*}
\normalsize
where $e \in S_q$ is an answer to the query, $e' \notin \sq$ represents a random negative sample, and $\gamma$ denotes the margin. $Z_q$ is an $L_2$ norm based scaling factor, which is introduced as a means to balance margin sensitivity between queries during training.
In the loss function, we use $k$ random negative samples and optimize the average.
We seek to maximize $\phi(q,e)$ for $e \in S_q$ and minimize $\phi(q, e')$ for $e' \in S_q$.

For the model inference, given a query $q$, \modelname~embeds it as $\bsq$ and rank all the entities by $\phi(q, \cdot)$.

\subsection{Theoretical Analysis} \label{sec:theory}
For \modelname, we present the following propositions with proof in Appendix \ref{ap:propproof}.
\begin{proposition}
Our conjunction operator $\conjop$ is commutative, associative, and satisfies conjunction elimination. 
\end{proposition}
\begin{proposition}
Our disjunction operator $\disjop$ is commutative, associative, and satisfies disjunction amplification. 
\end{proposition}
\begin{proposition}
Our negation operator $\negop$ is involutory and satisfies non-contradiction.
\end{proposition}

%% file: writeup/3-preliminaries.tex
\input{tables/logiclaws}

\input{tables/baselines}

\section{Preliminaries}
A knowledge graph (KG) consists of a set of triples $\langle e_s, r, e_o \rangle$, with $e_s, e_o \in \mathcal{E}$ (the set of entities) denoting the subject and object entities respectively and $r \in \mathcal{R}$ (the set of relations) denoting the relation between $e_s$ and $e_o$.
Without loss of generality, a KG can be represented as a First-Order Logic (FOL) Knowledge Base, where each triple $\langle e_s, r, e_o \rangle$ denotes an atomic formula $r(e_s, e_o)$, 
with $r \in \mathcal{R}$ denoting a binary predicate and $e_s, e_o \in \mathcal{E}$ as its arguments.

We aim to answer FOL queries expressed with existential quantification ($\exists$), conjunction ($\land$), disjunction($\lor$), and negation ($\neg$). 
The disjunctive normal form (DNF) of an FOL query $q$ is defined as follows:
\small
\begin{equation*}
    q[V_?] \triangleq V_?:
    \exists V_1, ..., V_k
    (v_{11} \land ... \land v_{1N_{1}}) \lor ... \lor (v_{M1} \land ... \land v_{MN_{M}}) 
\end{equation*} 
\normalsize
where $V_?$ is the \emph{target variable} of the query, and $V_1, ..., V_K$ denote the bound variable nodes. 
Each $v_{mn}$ ($m=1,...,M, n=1,...,N_m$) represents a literal, i.e., a logical atom or the negation of a logical atom: 
\begin{small}
\begin{equation*}
    v_{mn}= 
        \begin{cases}
            r(e, V) & r \in \mathcal{R}, e \in \mathcal{E}, V \in \{V_?, V_1, ..., V_k\} \\
            \neg r(e, V) & r \in \mathcal{R}, e \in \mathcal{E}, V \in \{V_?, V_1, ..., V_k\}\\ 
            r(V, V') & r \in \mathcal{R}, V \in \{V_1, ..., V_k\} \\
                     &  V' \in \{V_?, V_1, ..., V_k\}, V \neq V' \\
            \neg r(V, V') & r \in \mathcal{R}, V \in \{V_1, ..., V_k\} \\
            &  V' \in \{V_?, V_1, ..., V_k\}, V \neq V' \\
        \end{cases}
\end{equation*}
\end{small}



The goal of answering the logical query $q$ is to find a set of entities
$\sq = \{ a | a \in \mathcal{E}, q[a]~\text{holds true}\}$,
where $q[a]$ is a logical formula that substitutes the query target variable $V_?$ with the entity $a$. 



A complex query can be considered as a combination of multiple sub-queries.
For example, the query
$q[V_?]=V_?:$ Compose(John Lennon, $V_?$) $\land$ Compose(Paul McCartney, $V_?$)
can be considered as $q_1 \land q_2$, where
\begin{align*}
    & q_1[V_?]=V_?:\text{Compose(John Lennon}, V_?) \\
    & q_2[V_?]=V_?:\text{\emph{Compose}(Paul McCartney}, V_?) .
\end{align*}
Formally, we have:
\begin{align*}
    S_{q_1 \land q_2} & = S_{q_1} \cap S_{q_1};\\ 
    S_{q_1 \lor q_2}  & = S_{q_1} \cup S_{q_1};\\ 
    S_{\neg q} & = S_{q}^{\complement}
\end{align*}
where $(\cdot)^{\complement}$ denotes set complement.

Notation wise, we use boldfaced notations $\be$ and $\bsq$ to represent the embedding for entity $e$ and the embedding for $S_q$, i.e., the answer entity set for query $q$, respectively. 
We use $\psi_1, \psi_2, \psi_3$ to denote logical formulae.


\subsection{Logic Laws and Model Properties}
The general idea of logical query embedding models is to recursively define the embedding of a query (e.g., $q_1 \land q_2$) based on logical operations on its sub-queries' embeddings (e.g., $q_1$ and $q_2$). These logical operations have to satisfy logic laws, which serve as additional constraints to learning-based query embedding models. 
Unfortunately, most existing query embedding models have (partially) neglected these laws, which result in inferior performance.

In this section, we study these logic laws shared by both classical logic and basic fuzzy logic \cite{fuzzytheorybook} and deduce several basic properties that the logical operators should possess. The logic laws and corresponding model properties are summarized in Table \ref{tab:axioms}.

\subsubsection{Axiomatic Systems of Logic}
Let $\mathcal{L}$ be the set of all the valid logic formulae under a logic system, and
$\psi_1, \psi_2, \psi_3 \in \mathcal{L}$ represent logical formulae. $I(\cdot)$ denotes the truth value of a logical formula. 
The semantics of Boolean Logic are defined by (i) the interpretation $I: \mathcal{L} \rightarrow \{0,1\}$,
(ii) the Modus Ponen inference rule ``$\text{from}~\psi_1~\text{and}~
\psi_1 \rightarrow \psi_2 ~\text{infer}~\psi_2$'', which characterizes logic implication ($\rightarrow$) as follows:
\begin{equation*}
    \psi_1 \rightarrow \psi_2 \quad \text{holds if and only if} \quad I(\psi_2) \geq I(\psi_1)
\end{equation*}
and (iii) a set of axioms written in Hilbert-style deductive systems
\cite{KlirFuzzyBook}.
Those axioms define other
logic connectives via logic implication ($\rightarrow$); for example, the following three axioms characterize the conjunction ($\land$) of Boolean logic \cite{chvalovsky2012independence}:
\begin{align*}
    & \psi_1 \wedge \psi_2 \rightarrow \psi_1 \\
    & \psi_1 \wedge \psi_2 \rightarrow \psi_2 \\
    & (\psi_3 \rightarrow \psi_1) \rightarrow ((\psi_3 \rightarrow \psi_2) \rightarrow (\psi_3 \rightarrow \psi_1 \wedge \psi_2))
\end{align*}
The first two axioms guarantee that the truth value of $\psi_1 \wedge \psi_2$ never exceeds the truth values of $\psi_1$ and $\psi_2$, and the last one enforces that 
$I(\psi_1 \wedge \psi_2) = 1$ if $I(\psi_1)=I(\psi_2)=1$.
The three axioms also imply commutativity and associativity of logical conjunction $\land$.
More discussions about the axiomatic systems can be found in Appendix \ref{ap:axioms}. 

\subsubsection{Model Properties} \label{sec:properties}
Let $\phi(q,e)$ be the embedding model scoring function estimating the probability that the entity $e$ can answer the query $q$. 
This means that $\phi(q,e)$ estimates the truth value $I(q[e])$, where $q[e]$ is a logical formula that uses $e$ to fill $q$.
For example,
given the query
$q=V_?: \emph{Compose}(\emph{John Lennon}, V_?)$ and the entity $e =\emph{``Let it Be''}$, 
$\phi(q,e)$ estimates the truth value of the logical formula
$\emph{Compose}(\emph{John Lennon}, \emph{Let it Be})$.
We can thus use logic laws to deduce reasonable properties that a query embedding model should possess. 
For instance, 
$\psi_1 \wedge \psi_2 \rightarrow \psi_1$ is an axiom that characterizes logic conjunction ($\land$), which enforces that $I(\psi_1 \land \psi_2) \leq I(\psi_1)$, and we accordingly expect the embedding model to satisfy $\phi(q_1 \land q_2, e) \leq \phi(q_1, e)$, i.e., an entity $e$ is less likely to satisfy $q_1\land q_2$ than $q_1$. 

Based on the axioms and deduced logic laws of classical logic \cite{FuzzyLogicAxioms}, we summarize a series of model properties that a logical query embedding model should possess in Table \ref{tab:axioms}.
The list is not exhaustive but indicative.

\begin{figure}[t]
    \centering
    \includegraphics[width=\linewidth]{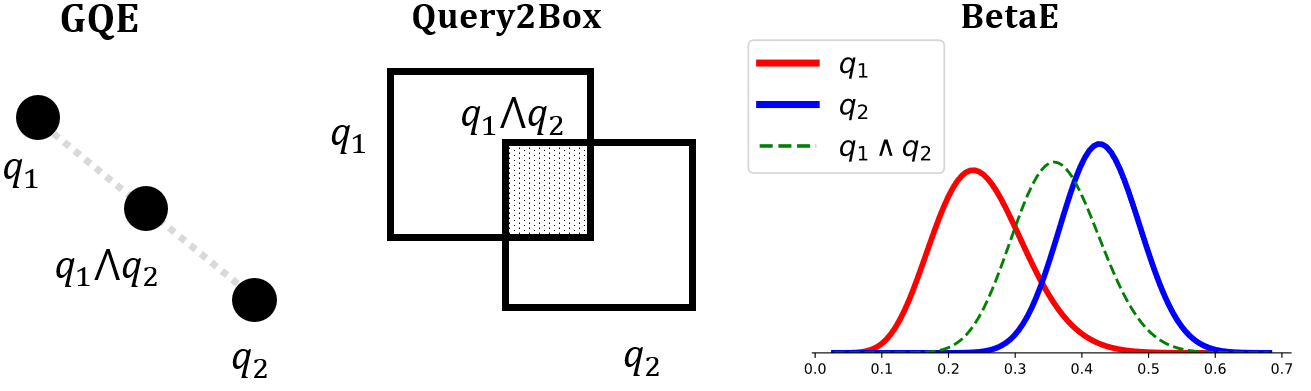}
    \caption{Illustration of query embeddings and embeddings of conjunctive queries in GQE, Query2Box, and BetaE. The conjunction operators takes embeddings of queries $q_1, q_2$ as input and produce an embedding for $q_1 \land q_2$.
    } 
    \label{fig:baselineconj}
\end{figure}

\subsection{Analysis of Prior Models on Model Properties} \label{sec:analysis}
This section examines three representative logical query embedding models, namely
GQE \cite{GQE}, Query2Box \cite{Query2Box}, and BetaE \cite{BetaE},
 regarding their capability of satisfying the properties in Table \ref{tab:axioms}. 
We summarize our findings in Table \ref{table:baselines}. 
GQE, Query2Box, BetaE represent queries as vectors, boxes (axis-aligned hyper-rectangles), and Beta distributions, respectively. The embedding-based logical operators transform embeddings of sub-queries into embeddings of the outcome query.
A brief summary of logical operators of these models are given in Appendix \ref{ap:baselineconj}.

\subsubsection{Conjunction ($\land$)}
Fig. \ref{fig:baselineconj} illustrates embedding-based conjunction operators of the three models, which take embeddings of queries $q_1, q_2$ as input and produce an embedding for $q_1 \land q_2$. 
GQE, Query2Box, and BetaE are purposely constructed to be permutation invariant \cite{GQE, Query2Box, BetaE}, and their conjunction operators all satisfy \emph{commutativity} (Law \myroman{2}).
The conjunction operators of GQE and BetaE do not satisfy \emph{associativity} (\myroman{3}) since they rely on the operation of averaging,
which is not associative.
GQE does not satisfy \emph{conjunction elimination} (\myroman{1}); 
for example, supposing that $\be = \frac{1}{2}(\bsq_1+\bsq_2)$, $\bsq_1 \neq \bsq_2$, we have $\phi(q_1 \land q_2, e)=\cos(\be, \frac{1}{2}(\bsq_1+\bsq_2)) >\cos(\be, \bsq_1)=  \phi(q_1, e)$. 
BetaE does not satisfy \emph{conjunction elimination} (\myroman{1}) for similar reasons.

\subsubsection{Disjunction ($\lor$)}
Previous works handle disjunction in two ways:
the \emph{Disjunctive Normal Form (DNF) rewriting} approach proposed by Query2Box \cite{Query2Box}, and the \emph{De Morgan's law (DM)} approach proposed by BetaE \cite{BetaE}.
The DNF rewriting method involves rewriting each query as a DNF to ensure that disjunction only appears in the last step, 
which enables the model to simply retain all input embeddings.
The model correspondingly cannot represent the disjunction result as a closed form; for example, the disjunction of two boxes remains two separate boxes instead of one \cite{Query2Box}. 
The DM approach uses De Morgan's law $\psi_1 \lor \psi_2 \equiv \neg(\neg \psi_1 \land \neg \psi_2)$ to compute the disjunctive query embedding,
which requires the model to have a conjunction operator and a negation operator. 
This approach advantageously
produces representation in a closed form, allowing disjunction to be performed at any step of the computation. The disadvantage is that if the negation operator does not work well, the error will be amplified and affect disjunction. The DM variant of BetaE, namely BetaE$_{\text{DM}}$, does not satisfy \emph{disjunction amplification} (\myroman{4})
since its negation operator violates \emph{non-contradiction} (\myroman{8}). 

\subsubsection{Negation}
To the best of our knowledge, BetaE is the only previous model that can handle negation.
BetaE has proved that its negation operator is \emph{involutory} (\myroman{7}) \cite{BetaE}. 
However, this operator lacks the \emph{non-contradiction} property (\myroman{8}), as 
for BetaE $\phi(\neg q, e)$ is not monotonically decreasing with regard to $\phi(q, e)$.
An illustration is given in Appendix \ref{sec:beta2fail}.

\subsection{Fuzzy Logic} \label{sec:fuzzylogic}
Fuzzy logic differs from Boolean logic by associating every logical formula with a truth value in $[0,1]$.
Fuzzy logic systems usually retain the axioms of Boolean logic,
which ensures that all the logical operation behaviors are consistent with Boolean logic when the truth values are $0$ or $1$. 
Different fuzzy logic systems add different axioms to define the logical operation behavior for the case when the truth value is in $(0,1)$ \cite{KlirFuzzyBook}. 
A $t$-norm $\top: [0,1] \times [0,1] \mapsto [0,1]$ represents generalized conjunction in fuzzy logic.
Prominent examples of \tnorms~include 
\godel~t-norm $\top_{\min}\{x,y\} = \min(x,y)$, 
product t-norm $\top_{\text{prod}}\{x,y\} = xy$, 
and \luka~\tnorm $\top_{\luka}(x,y) = \max\{0, x+y-1\}$, for $x,y\in[0,1]$. 
Any other continuous \tnorm~can be described as an ordinal sum of these three basic ones \cite{KlementTNormBook}.
Analogously, $t$-conorm are dual to $t$-norms for disjunction in fuzzy logic \textendash~given a $t$-norm $\top$, the t-conorm is defined as $\bot(x,y) = 1 - \top(1-x, 1-y)$ based on De Morgan's law and the negator $n(x)=1-x$ for $x,y \in [0,1]$
\cite{KlementTNormBook}. 
This technique has inspired numerous subsequent works. For example,
CQD \cite{CQD} uses t-norms and t-conorms to rank entities for query answering on KGs.


%% file: tables/logiclaws.tex
\begin{table*}[t]
\centering
\setlength{\tabcolsep}{10pt}
\begin{tabular}{cc|l|l}
\toprule
     & & Logic Law & Model Property \\
\midrule
\multirow{7}{*}{$\land$} 
    & \multirow{3}{*}{\myroman{1}} & Conjunction Elimination  & \\
    &  & $\psi_1 \wedge \psi_2 \rightarrow \psi_1$
    & $\phi(q_1 \land q_2, e) \leq \phi(q_1, e) $ \\
    
    & & $\psi_1 \wedge \psi_2 \rightarrow \psi_2$
    & $\phi(q_1 \land q_2, e) \leq \phi(q_2, e) $ \\
    \cmidrule(lr){2-4}
    
    & \multirow{2}{*}{\myroman{2}} & Commutativity & \\
    &  & $\psi_1 \land \psi_2 \leftrightarrow \psi_2 \land \psi_1$
        & $\phi((q_1 \land q_2), e) = \phi((q_2 \land q_1), e)  $ \\
    \cmidrule(lr){2-4}
    
    & \multirow{2}{*}{\myroman{3}} &  Associativity & \\
    & & $(\psi_1 \land \psi_2) \land \psi_3 \leftrightarrow \psi_1 \land (\psi_2 \land \psi_3)$
        & $\phi((q_1 \land q_2) \land q_3, e)= \phi(q_1 \land (q_2 \land q_3), e)$ \\
\midrule

\multirow{7}{*}{$\lor$} 
    & \multirow{3}{*}{\myroman{4}} & Disjunction Amplification  & \\
    &  & $\psi_1 \rightarrow \psi_1 \vee \psi_2$
    & $\phi(q_1, e) \leq \phi(q_1 \lor q_2, e) $ \\
    
    & & $\psi_2 \rightarrow \psi_1 \vee \psi_2$
    & $\phi(q_2, e) \leq \phi(q_1 \lor q_2, e) $ \\
    \cmidrule(lr){2-4}

    & \multirow{2}{*}{\myroman{5}} & Commutativity & \\
    &  & $\psi_1 \lor \psi_2 \leftrightarrow \psi_2 \lor \psi_1$
        & $\phi((q_1 \lor q_2), e) = \phi((q_2 \lor q_1), e)  $  \\
    \cmidrule(lr){2-4}

    &  \multirow{2}{*}{\myroman{6}} & Associativity & \\
    &  & $(\psi_1 \lor \psi_2) \lor \psi_3 \leftrightarrow \psi_1 \lor (\psi_2 \lor \psi_3)$
        & $\phi((q_1 \lor q_2) \lor q_3, e) = \phi(q_1 \lor (q_2 \lor q_3), e) $ \\
\midrule

\multirow{4}{*}{$\neg$} 
    &  \multirow{2}{*}{\myroman{7}} & Involution & \\
    &  & $\neg \neg \psi_1 \rightarrow \psi_1$ 
        & $\phi(q, e) = \phi(\neg \neg q, e)$ \\
    \cmidrule(lr){2-4}
    
    &  \multirow{2}{*}{\myroman{8}} & Non-Contradiction & \\
    & & $\psi_1 \land \neg \psi_1 \rightarrow \overline{0}$
        & $\phi(q, e) \uparrow \quad \Rightarrow \quad \phi(\neg q, e) \downarrow $\\
\bottomrule  
\end{tabular}
\caption{
Here we list eight logic laws (I - VIII) 
from classical logic 
\cite{fuzzytheorybook} and give the corresponding properties that a query embedding model should possess.
$\psi_1, \psi_2, \psi_3$ represent logical formulae.
$\phi$ denotes the scoring function that
estimates the probability that the entity $e$ can answer the query $q$. 
$\phi(q, e) \uparrow \Rightarrow \phi(\neg q, e) \downarrow $ means $\phi(\neg q, e)$ is monotonically decreasing with regard to $\phi(q, e)$. 
}
\label{tab:axioms}
\end{table*}

%% file: tables/baselines.tex
\begin{table*}[t]
\footnotesize
\centering
\setlength{\tabcolsep}{2pt}
\begin{tabular}{cccccrccccccccc}

\toprule
  &     \multicolumn{4}{c}{$\land$ } &   \multicolumn{4}{c}{$\lor$}&  \multicolumn{3}{c}{$\neg$}\\
  
  \cmidrule(lr){2-5} \cmidrule(lr){6-9} \cmidrule(lr){10-12} 

  &           Expressivity (Closed) &       \emph{Com.} &       \emph{Asso.} &       \emph{Elim.} 
  &           Expressivity (Closed) &       \emph{Com.} &       \emph{Asso.} &       \emph{Ampli.}
  &           Expressivity (Closed)&  \emph{Inv.} &       \emph{Non-Contra.} \\
\midrule

GQE 
    &  \checkmark (\checkmark) &  \checkmark &  \xmark &      \xmark 
    &      \checkmark (\xmark) &  \checkmark &  \checkmark &      \checkmark 
    &                   \xmark &           N/A &           N/A \\
    \midrule
Query2Box
    &  \checkmark (\checkmark) &  \checkmark &  \checkmark &  \checkmark 
    &      \checkmark (\xmark) &  \checkmark &  \checkmark &  \checkmark 
    &                   \xmark &           N/A &           N/A \\
    \midrule

\multirow{2}{*}{BetaE}
  &  \multirow{2}{*}{\checkmark (\checkmark)} &  \multirow{2}{*}{\checkmark} & \multirow{2}{*}{\xmark}  & \multirow{2}{*}{\xmark} 
  &  (i) DNF \checkmark (\xmark) &  \checkmark &  \checkmark &  \checkmark 
  &  \multirow{2}{*}{\checkmark (\checkmark)} &  \multirow{2}{*}{\checkmark}  &  \multirow{2}{*}{\xmark}
\\
  &   &  &  &  
  &  (ii) DM \checkmark (\checkmark) &  \checkmark &  \checkmark &  \xmark 
  &  &   & 
\\
\midrule

    
    

 $\modelname$ &  \checkmark (\checkmark) &  \checkmark &  \checkmark &  \checkmark &  \checkmark (\checkmark) &  \checkmark &  \checkmark &  \checkmark &  \checkmark (\checkmark) &  \checkmark &  \checkmark \\
\bottomrule

\end{tabular}
\caption{Comparisons of different models regarding the properties of logical operations.
\emph{Expressivity} indicates whether the model can handle such logical operations, and \emph{closed} indicates whether the embedding is in a closed form. \emph{Commu., Asso., Elim., Ampli., Inv.} and \emph{Non-contra.} stand for commutativity, associativity, conjunction elimination, disjunction amplification, involution, and non-contradiction respectively.
}
\label{table:baselines}
\end{table*}

%% file: writeup/5-experiments.tex
\section{Experiments}

\input{tables/results-main}
\input{tables/results-projectiononly}





In this section, we evaluate the ability of \modelname~ to answer complex FOL queries over incomplete KGs.


\subsection{Evaluation Setup}

\paragraph{Datasets}
We evaluate our model on two benchmark datasets provided by \cite{BetaE}, which contain 14 types of logical queries on 
FB15k-237 \cite{FB15k237} and NELL995 \cite{DeepPath} respectively. 
The 14 types of query structures in the datasets are shown in Fig. \ref{fig:query_structure}.
Note that these datasets provided by BetaE \cite{BetaE} are an improved and expanded version of the datasets provided by Query2Box \cite{Query2Box}.
Compared to the earlier version, the new datasets \citet{BetaE} contain 5 new types of queries that involve negation. 
The validation/test set of the original 9 query types are regenerated to ensure that the number of answers per query is not excessive, making this task more challenging. 
In the new datasets, $10$ query structures are used for both training and evaluation: $1p, 2p, 3p, 2i, 3i, 2in, 3in, inp, pni, pin$. $4$ query structures ($ip, pi, 2u, up$) are not used for training but only included in evaluation in order to evaluate the model's generalization ability of answering queries with logical structures that the model has never seen during training.
We exclude FB15k \cite{TransE} since this dataset suffers from major test leakage \cite{FB15k237}.
Statistics about the datasets are summarized in Appendix \ref{ap:datasets}.

\paragraph{Evaluation Protocol}
We follow the evaluation protocol in \cite{BetaE}.
To evaluate the model's generalization capability over incomplete KGs, the datasets are masked out so that each validation/test query answer pair involves imputing at least one missing edge.
For each answer of a test query, we use the Mean Reciprocal Rank (MRR) as the major evaluation metric.
We use the \emph{filtered} setting \cite{TransE} and filter out other correct answers from ranking before calculating the MRR.

\paragraph{Baselines and Model Configurations} 
We consider three logical query embedding baselines for answering complex logical queries on KGs:
GQE \cite{GQE}, Query2Box \cite{Query2Box}, and BetaE \cite{BetaE}. We also compare with one recent state-of-the-art query optimization model CQD \cite{CQD}.
For GQE, Query2Box, and BetaE we use implementation provided by \cite{BetaE} \footnote{\url{https://github.com/snap-stanford/KGReasoning}}.
For BetaE and CQD, we compare with the model variant that generally provides better performance, namely BetaE$_{\text{DNF}}$ and CQD-BEAM.
CQD cannot process complex logical queries during training and is thus trained with KG edges.
To the best of our knowledge, BetaE is the only available baseline that can handle negation. Therefore, for GQE, Query2Box, and CQD, we compare with them only on EPFO queries (queries with $\exists, \land, \lor$ and without negation).

For \modelname, we report results using the logic system that provide the best average MRR on the validation set. we use AdamW \cite{AdamW} as the optimizer. Training terminates with early stopping based on the average MRR on the validation set with a patience of 15k steps. 
We repeat each experiment three times with different random seeds and report the average results.
Hyperparameters and more experimental details are given in Appendix \ref{ap:experiments}.

\subsection{Main Results: Trained with FOL queries}
We first test the ability of \modelname to model arbitrary FOL queries when complex logical queries are available for training.
Results are reported in Table~\ref{tab:main}.
\subsubsection{Comparison with Query Embedding}
As shown in Table~\ref{tab:main}, \modelname
consistently outperforms all the logical query embedding baselines.
For EPFO queries, 
\modelname improves the average MRR of best baseline BetaE \cite{BetaE} by 3.3\% (ca. 15\% relative improvement) on FB15k-237 and 4.7\% (ca. 19\% relative improvement) on NELL995.
For queries with negation, 
\modelname~significantly outperforms the only available baseline BetaE. 
On average, \modelname~improves the MRR by 3.0\% (54\% relatively) on FB15k-237 and 2.1\% (36\% relatively) on NELL995 for queries containing negation.
We hypothesize that this significant enhancement comes from the principled design of our negation operator that satisfies the axioms, while BetaE fails to satisfy the non-contradiction property. 




\paragraph{Comparison with Query Optimization: CQD} We next compare \modelname with a recent query optimization baseline, CQD~\cite{CQD} on EPFO queries.
On average, \modelname provides 2.5\% and 0.9\% absolute improvement in MRR on FB15k-237 and NELL995 respectively. 
It is worth noting that \modelname outperforms CQD on most complex query structures on NELL995 even with slightly worse $1p$ query answering performance.
We hypothesize that the $1p$ query answering performance difference on NELL995 comes from the differenct abilities of different relation projection/link prediction models to encode sparse knowledge graphs.

A major motivation for learning logical query embedding is its high inference efficiency.
We compare with CQD with regard to the time for answering a query.
On a $\text{NVIDIA}^{\circledR}$ GP102 TITAN Xp (12GB),
the average time for CQD to answer a FOL query on FB15k-237 is 13.9 ms (milliseconds), while FuzzQE takes only 0.3 ms.
On NELL995, where the number of entities is 4 times the number in FB15k-237, the average time for CQD is 68.1 ms, whereas FuzzQE needs only 0.4 ms. CQD takes 170 times longer than FuzzQE.
The reason is that CQD is required to score all the entities for each subquery to obtain the top-$k$ candidates for beam search.


\subsection{Trained with only Link Prediction}
This experiment tests the ability of the model to generalize to arbitrary complex logical queries when it is trained with only the link prediction task.
To evaluate it, we train \modelname and other logical query embedding models using only KG edges (i.e., \emph{1p} queries).
For baseline models GQE, Query2Box, and BetaE, we adapt them following the experiment settings of the Q2B-AVG-1P model discussed in \cite{Query2Box}. 
Specifically, 
we set all the sub-query weights to $1.0$ for this experiment.

As shown in Table \ref{table:1p}, 
\modelname~is able to generalize to complex logical queries of new query structures even if it is trained on link prediction and provides significantly better performance than baseline models. 
Compared to the best baseline,
\modelname improves the average MRR by 3.6\% (20\% relatively) for EPFO queries on FB15k-237 and 5.4\% (26\% relatively) on NELL995. 
Regarding queries with negation, our model drastically outperforms the only available baseline BetaE across datasets.
In addition, compared with the ones trained with complex FOL queries (in Table~\ref{tab:main}),
It is worth nothing that \modelname trained with only link prediction can outperform BetaE models that are trained with extra complex logical queries in terms of average MRR (in Table~\ref{tab:main}).
This demonstrates the superiority of the logical operators in \modelname, which are designed in a principled and learning-free manner.
Meanwhile, \modelname can still take advantage of additional complex queries as training samples to enhance entity embeddings.

%% file: tables/results-main.tex
\begin{table*}[t]
\footnotesize
\centering
\setlength{\tabcolsep}{3pt}
\begin{tabular}{c|c|cc|ccccccccc|ccccc}
\toprule
Type of Model &     Model & Avg$_{\text{EPFO}}$ & Avg$_{\text{Neg}}$ &   1p &   2p &   3p &   2i &   3i &   pi &   ip &   2u &   up & 2in &  3in &  inp & pin & pni \\
\midrule
\multicolumn{18}{c}{FB15k-237} \\ \midrule 
\multirow{4}{*}{{Query Embedding}} &   GQE &    16.3 &     N/A & 35.0 &  7.2 &  5.3 & 23.3 & 34.6 & 16.5 & 10.7 &  8.2 &  5.7 & N/A &  N/A &  N/A & N/A & N/A \\
            &      Query2Box &    20.1 &     N/A & 40.6 &  9.4 &  6.8 & 29.5 & 42.3 & 21.2 & 12.6 & 11.3 &  7.6 & N/A &  N/A &  N/A & N/A & N/A \\
            &    BetaE &    20.9 &     5.5 & 39.0 & 10.9 & 10.0 & 28.8 & 42.5 & 22.4 & 12.6 & 12.4 &  9.7 & 5.1 &  7.9 &  7.4 & 3.5 & 3.4 \\
 &  \modelname &\textbf{24.2}&\textbf{8.5}& 42.2&\textbf{13.3}&\textbf{10.2}&\textbf{33.0}&\textbf{47.3}&\textbf{26.2}&\textbf{18.9}&\textbf{15.6}&\textbf{10.8}&\textbf{9.7}&\textbf{12.6}&\textbf{7.8}&\textbf{5.8}&\textbf{6.6}\\ \midrule
{Query Optimization} &     CQD &	21.7 & N/A & \textbf{46.3} & 	9.9 &	5.9	& 31.7 &	41.3 &	21.8 &	15.8 & 	14.2 &	8.6 & N/A &  N/A &  N/A & N/A & N/A \\
\midrule \multicolumn{18}{c}{NELL995} \\ \midrule 
\multirow{4}{*}{{Query Embedding}} &     GQE &    18.6 &     N/A & 32.8 & 11.9 &  9.6 & 27.5 & 35.2 & 18.4 & 14.4 &  8.5 &  8.8 & N/A &  N/A &  N/A & N/A & N/A \\
           &     Query2Box &    22.9 &     N/A & 42.2 & 14.0 & 11.2 & 33.3 & 44.5 & 22.4 & 16.8 & 11.3 & 10.3 & N/A &  N/A &  N/A & N/A & N/A \\
           &     BetaE &    24.6 &     5.9 & 53.0 & 13.0 & 11.4 & 37.6 & 47.5 & 24.1 & 14.3 & 12.2 &  8.5 & 5.1 &  7.8 & 10.0 & 3.1 & 3.5 \\
 &   \modelname & \textbf{29.3} & \textbf{8.0} & 58.1 & \textbf{19.3} & \textbf{15.7}  & 39.8 & \textbf{50.3} & \textbf{28.1} & \textbf{21.8} & \textbf{17.3} & \textbf{13.7} &\textbf{8.3} & \textbf{10.2} & \textbf{11.5} & \textbf{4.6} & \textbf{5.4} \\ \midrule
 {Query Optimization} &     CQD & 28.4 & N/A &		\textbf{60.0} &	16.5 &	10.4 &	\textbf{40.4} &	49.6 &	27.6 &	20.8 &	16.8 &	12.6 & N/A &  N/A &  N/A & N/A & N/A \\
\bottomrule
\end{tabular}
\caption{MRR results (\%) on answering FOL queries. 
We report MRR results (\%) on test FOL queries. 
Avg$_{\text{EPFO}}$ and Avg$_{\text{Neg}}$ denote the average MRR on EPFO queries (queries with $\exists, \land, \lor$ and without negation) and queries containing negation respectively. 
Results of GQE, Query2Box, and BetaE are taken from \cite{BetaE}.
}
\label{tab:main}
\end{table*}

%% file: tables/results-projectiononly.tex
\begin{table*}[t]
\footnotesize
\centering
\setlength{\tabcolsep}{5pt}
\begin{tabular}{c|cc|ccccccccc|ccccc}
\toprule
      Model & Avg$_{\text{EPFO}}$ & Avg$_{\text{Neg}}$ &   1p &   2p &   3p &   2i &   3i &   pi &   ip &   2u &   up &    2in &    3in &    inp &    pin &    pni \\
\midrule \multicolumn{17}{c}{FB15k-237} \\ \midrule 
    GQE  &    17.7 &  N/A & 41.6 &  7.9 &  5.4 & 25.0 & 33.6 & 16.3 & 10.9 & 11.9 &  6.2 & N/A & N/A & N/A & N/A & N/A \\
    Query2Box  &    18.2 &  N/A & 42.6 &  6.9 &  4.7 & 27.3 & 36.8 & 17.5 & 11.1 & 11.7 &  5.5 & N/A & N/A & N/A & N/A & N/A \\
    BetaE  &    15.8 &     0.5 & 37.7 &  5.6 &  4.4 & 23.3 & 34.5 & 15.1 &  7.8 &  9.5 &  4.5 &    0.1 &    1.1 &    0.8 &    0.1 &    0.2 \\
 \modelname &\textbf{21.8}&\textbf{6.6}&\textbf{44.0}&\textbf{10.8}&\textbf{8.6}&\textbf{32.3}&\textbf{41.4}&\textbf{22.7}&\textbf{15.1}&\textbf{13.5}&\textbf{8.7}&\textbf{7.7}&\textbf{9.5}&\textbf{7.0}&\textbf{4.1}&\textbf{4.7}\\
\midrule \multicolumn{17}{c}{NELL995} \\ \midrule 
    GQE  &   21.7 &  N/A & 47.2 & 12.7 &  9.3 & 30.6 & 37.0 & 20.6 & 16.1 & 12.6 & 9.6 & N/A & N/A & N/A & N/A & N/A \\
    Query2Box  &    21.6 &  N/A & 47.6 & 12.5 &  8.7 & 30.7 & 36.5 & 20.5 & 16.0 & 12.7 &  9.6 & N/A & N/A & N/A & N/A & N/A \\
    BetaE  &    19.0 &     0.4 & 53.1 &  6.0 &  3.9 & 32.0 & 37.7 & 15.8 &  8.5 & 10.1 &  3.5 &    0.1 &    1.4 &    0.1 &    0.1 &    0.1 \\
\modelname& \textbf{27.1} & \textbf{7.3} & \textbf{57.6} & \textbf{17.2} & \textbf{13.3} & \textbf{38.2} & \textbf{41.5} & \textbf{27.0} & \textbf{19.4} & \textbf{16.9} & \textbf{12.7} & \textbf{9.1} & \textbf{8.3} & \textbf{8.9} & \textbf{4.4} & \textbf{5.6} \\
\bottomrule
\end{tabular}
\caption{MRR results (\%) of logical query embedding models that are trained with only link prediction. This task tests the ability of the model to generalize to arbitrary complex logical queries, when no complex logical query data is available for training.
Avg$_{\text{EPFO}}$ and Avg$_{\text{Neg}}$ denote the average MRR on EPFO ($\exists, \land, \lor$) queries and queries containing negation respectively. 
}
\label{table:1p}
\end{table*}

%% file: writeup/6-conclusion.tex
\section{Conclusion}
We propose a novel logical query embedding framework \modelname~for answering complex logical queries on KGs. 
Our model \modelname~borrows operations from fuzzy logic and implements logical operators in a principled and learning-free manner. 
Extensive experiments show the promising capability of \modelname~on answering logical queries on KGs.
The results are encouraging and suggest various extensions, 
including introducing logical rules into embedding learning, 
as well as studying the potential use of predicate fuzzy logic systems and other deeper transformation architectures.
Future research could also use the defined logical operators to incorporate logical rules to enhance reasoning on KGs. 
Furthermore, we are interested in jointly learning embeddings for logical queries, natural language questions, entity labels to enhance question answering on KGs.

%% file: writeup/7-0-appendix.tex
\newpage
\clearpage
\appendix
\onecolumn

\section*{Appendix}

\input{writeup/7-A-proposition-proof}


\input{writeup/7-C-axiomaticsystem}
\input{writeup/7-D-baselineoperators}

\input{writeup/7-E-dataset}

\input{writeup/7-F-experiment.tex}

\input{writeup/7-G-fuzzylogic.tex}

\input{writeup/7-H-time.tex}

\newpage
\clearpage

%% file: writeup/7-A-proposition-proof.tex
\section{Proof of propositions} \label{ap:propproof}
We hereby provide proof of propositions for FuzzQE using product logic.
The same can be proved for \godel~logic.

\subsection{Proof of Proposition 1} \label{ap:prop1}
\subsubsection{Commutativity $\phi(q_1 \land q_2, e) = \phi(q_2 \land q_1, e) $} 
\begin{proof}

We have
$\conjop(\bsq_1, \bsq_2) = q_1 \circ q_2 =  q_2 \circ q_1 = \conjop(\bsq_2, \bsq_1)$
where $\circ$ denotes element-wise multiplication.\\
Therefore,
$\phi(q_1 \land q_2, e) = \be ^ \intercal \conjop(\bsq_1, \bsq_2) = \be ^ \intercal \conjop(\bsq_2, \bsq_1) = \phi(q_2 \land q_1, e)$.
\end{proof}

\subsubsection{Associativity
$\phi((q_1 \land q_2) \land q_3, e) =  \phi(q_1 \land (q_2 \land q_3), e)$ }
\begin{proof}
Since
$\conjop(\conjop(\bsq_1, \bsq_2)), \bsq_3) = q_1 \circ q_2 \circ q_3 = \conjop (\bsq_1,  \conjop(\bsq_2,  \bsq_3))$,
we have
\begin{align*}
 & \phi((q_1 \lor q_2) \lor q_3, e) \\
=& \be^\intercal \conjop(\conjop(\bsq_1, \bsq_2)), \bsq_3) \\
= & \be^\intercal \conjop (\bsq_1,  \conjop(\bsq_2,  \bsq_3))\\
=& \phi(q_1 \lor (q_2 \lor q_3), e) 
\end{align*}
\end{proof}

\subsubsection{Conjunction elimination 
$\phi(q_1 \land q_2, e) \leq \phi(q_1, e) $, \quad
$\phi(q_1 \land q_2, e) \leq \phi(q_2, e) $}
\begin{proof}
$\phi(q_1 \land q_2, e) \leq \phi(q_1, e)$ can be proved by
\begin{align*}
     & \phi(q_1 \land q_2, e)\\
    =& \be^\intercal \conjop(\bsq_1, \bsq_2)\\
    =& \be^\intercal (\bsq_1 \circ \bsq_2)\\
    =& \sum_{i=1}^d \be_i \bsq_{1_i} \bsq_{2_i}\\
    \leq & \sum_{i=1}^d \be_i \bsq_{1_i}\\
    =& \phi(q_1, e)
\end{align*}
$\phi(q_1 \land q_2, e) \leq \phi(q_2, e)$ can be proved similarly.
\end{proof}

\subsection{Proof of Proposition 2} \label{ap:prop1}
\subsubsection{Commutativity $\phi(q_1 \lor q_2, e) = \phi(q_2 \lor q_1, e) $} 
\begin{proof}

We have
$\disjop(\bsq_1, \bsq_2) 
= \bsq_1 + \bsq_2 - \bsq_1 \circ \bsq_2 
=  \bsq_2 + \bsq_1 - \bsq_2 \circ \bsq_1 = \disjop(\bsq_2, \bsq_1)$.\\
Therefore,
$\phi(q_1 \lor q_2, e) = \be ^ \intercal \disjop(\bsq_1, \bsq_2) = \be ^ \intercal \disjop(\bsq_2, \bsq_1) = \phi(q_2 \lor q_1, e)$.
\end{proof}

\subsubsection{Associativity
$\phi((q_1 \land q_2) \land q_3, e) =  \phi(q_1 \land (q_2 \land q_3), e)$ }
\begin{proof}
\begin{align*}
    & \disjop(\disjop(\bsq_1, \bsq_2)), \bsq_3) \\
=& \disjop(\bsq_1 + \bsq_2 - \bsq_1 \circ \bsq_2, \bsq_3) \\
=& (\bsq_1 + \bsq_2 - \bsq_1 \circ \bsq_2) + \bsq_3 - (\bsq_1 + \bsq_2 - \bsq_1 \circ \bsq_2) \circ \bsq_ 3 \\
=& \bsq_1 + \bsq_2 + \bsq_3 - \bsq_1 \circ \bsq_2 - \bsq_1 \circ \bsq_3 -\bsq_2 \circ \bsq_3 + \bsq_1 \circ \bsq_2 \circ \bsq_3 \\
=& \disjop(\bsq_1, \disjop(\bsq_2, \bsq_3))
\end{align*}
Therefore
\begin{align*}
& \phi((q_1 \lor q_2) \lor q_3, e) \\
= & \be^\intercal \disjop(\disjop(\bsq_1, \bsq_2)), \bsq_3) \\
= & \be^\intercal \disjop (\bsq_1,  \disjop(\bsq_2,  \bsq_3))\\
= & \phi(q_1 \lor (q_2 \lor q_3), e) 
\end{align*}
\end{proof}


\subsubsection{Disjunction amplification 
$\phi(q_1 \lor q_2, e) \geq \phi(q_1, e) $, \quad
$\phi(q_1 \lor q_2, e) \geq \phi(q_2, e) $}
\begin{proof}
$\phi(q_1 \lor q_2, e) \geq \phi(q_1, e)$ can be proved by
\begin{align*}
& \phi(q_1 \lor q_2, e) \\
=& \be^\intercal \disjop(\bsq_1, \bsq_2)\\
=& \be^\intercal (\bsq_1 + \bsq_2 - \bsq_1 \circ \bsq_2) \\
=& \sum_{i=1}^d \be_i (\bsq_{1_i} + \bsq_{2_i} - \bsq_{1_i} \bsq_{2_i})\\
=& \sum_{i=1}^d \be_i \bsq_{1_i} + \be_i \bsq_{2_i} (1- \bsq_{1_i})\\
\geq & \sum_{i=1}^d \be_i \bsq_{1_i}\\
=& \phi(q_1, e)
\end{align*}
$\phi(q_1 \lor q_2, e) \geq \phi(q_2, e)$ can be proved similarly.
\end{proof}

\subsection{Proof of Proposition 3} \label{ap:prop2}
\subsubsection{Involution $\phi(q, e) = \phi(\neg \neg q, e)$}
\begin{proof}
\begin{align*}
    \negop(\negop(q)) = \mathbf{1} - (\mathbf{1} - \bsq) = \bsq
\end{align*}
Therefore $\phi(\neg \neg q, e) = \be^\intercal \negop(\negop(\bsq)) = \phi(q,e)$
\end{proof}

\subsubsection{Non-contradiction
$\phi(q, e) \uparrow \Rightarrow \phi(\neg q, e) \downarrow $\
}
\begin{proof}
The \luka~negation $c(x)=1-x$ is monotonically decreasing with regard to $x$.
Therefore, $\phi(\neg q, e)$ is monotonically decreasing with regard to $\phi(q, e)$.
\end{proof}


%% file: writeup/7-C-axiomaticsystem.tex
\newpage
\clearpage

\section{Axiomatic systems of logic} \label{ap:axioms}

Axiomatic systems of logic consist of a set of axioms and the Modus Ponen inference rule:
    $\text{from}~\varphi~\text{and}~
    \varphi \rightarrow \psi
    ~\text{infer}~\psi$.
Implication $\rightarrow$ is defined as \emph{$\varphi \rightarrow \psi$ holds if the truth value of $\psi$ is larger than or equal to $\varphi$}.
In Table \ref{tab:semantics}, we compare the semantics of classical logic and product logic and show that product logic operations are fully compatible with classical logic.
In Table \ref{tab:fullaxioms}, we provide the list of axioms written in Hilbert-style deductive system for classical logic, and three prominent fuzzy logic systems: \luka~logic, \godel~logic, and product logic \cite{KlementTNormBook}.
We also provide some of the derivable logic laws. Interested readers are referred to \cite{fuzzytheorybook} for proofs.

\input{tables/semantics-product}


\input{tables/axioms-full}

%% file: tables/semantics-product.tex
\begin{table}[h]
\centering
\caption{
Semantics of classical logic and product logic.
$F_{mL}$ denote all valid logic formulae under the logic system, and 
$\varphi, \psi \in F_{mL}$ are logical formulae.
$I(\cdot)$~denotes the truth value of a logical formula.
}
\begin{tabular}{lll}
\toprule
 &  Classical Logic & Product Logic \\
\midrule
Interpretation $I$
    & $I: F_{mL} \rightarrow \{0,1\}$
    & $I: F_{mL} \rightarrow [0,1]$
    \\
\midrule
$I(\varphi \land \psi)$
    & $I(\varphi) I(\psi)$
    & $I(\varphi) I(\psi)$
    \\
\midrule
$I(\varphi \lor \psi)$
    & $I(\varphi)+I(\psi) - I(\varphi)I(\psi) $
    & $I(\varphi)+I(\psi) - I(\varphi)I(\psi) $
    \\
\midrule
$I(\varphi \rightarrow \psi)$
    & 
        $\left\{
             \begin{array}{lr}
             1, \quad \text{if}~I(\varphi) \leq I(\psi)  \\
             I(\psi), \quad \text{otherwise}
             \end{array}
        \right.
        $
    & 
        $\left\{
             \begin{array}{lr}
             1, \quad \text{if} I(\varphi) \leq I(\psi)  \\
             I(\psi), \quad \text{otherwise}
             \end{array}
        \right.
        $
    \\
    
    



\bottomrule  
\end{tabular}
\label{tab:semantics}
\end{table}

%% file: tables/axioms-full.tex
\begin{table}[h]
\footnotesize
\centering
\caption{Axioms and derivable logic laws of classical logic, basic fuzzy logic, and three prominent fuzzy logic systems that are based on on basic fuzzy logic: \luka~logic, \godel~logic, and product logic \cite{KlementTNormBook}.
\textbullet~denotes that the formula is in the minimal axiomatic system \cite{chvalovsky2012independence}, while
\textopenbullet~means the logic law could be derived from the minimal axiomatic system.
\emph{EFQ} stands for \emph{Ex falso quodlibet}, which is Latin for \emph{from falsehood, anything}.
\emph{Strong conjunction} of fuzzy logic is usually defined by $t$-norm (See Appendix \ref{ap:tnorm}) and denoted by $\&$ in literature. Here, to make it easier to compare with classical logic, we uniformly use $\land$ in the axioms. 
}
\begin{tabular}{llccccc}
\toprule
     & Axiom / Logic Law & Classical Logic & Basic Fuzzy Logic & \luka & Gödel & Product \\
\midrule
Transitivity
    & $(\varphi \rightarrow \chi) \rightarrow((\psi \rightarrow \chi) \rightarrow(\varphi \rightarrow \chi))$
    & \textbullet
    & \textbullet
    & \textbullet 
    & \textbullet 
    & \textbullet
    \\
\midrule

Weakening
    & $\varphi \rightarrow(\psi \rightarrow \varphi)$
    & \textbullet
    & \textbullet
    & \textbullet 
    & \textbullet 
    & \textbullet
    \\
\midrule

Exchange
    & $(\varphi \rightarrow(\psi \rightarrow \chi)) \rightarrow(\psi \rightarrow(\varphi \rightarrow \chi))$
    & \textbullet
    & \textbullet
    & \textbullet 
    & \textbullet 
    & \textbullet
    \\
\midrule

$\land$(\myroman{1}) 
    & $\varphi \wedge \psi \rightarrow \varphi$
    & \textbullet
    & \textbullet 
    & \textbullet 
    & \textbullet 
    & \textbullet
    \\
\midrule
    
$\land$(\myroman{2}) 
    & $\varphi \wedge \psi \rightarrow \psi$
    & \textbullet
    & \textbullet
    & \textbullet 
    & \textbullet 
    & \textbullet
    \\
\midrule
    
$\land$(\myroman{3})
    & $(\chi \rightarrow \varphi) \rightarrow ((\chi \rightarrow \psi) \rightarrow (\chi \rightarrow \varphi \wedge \psi))$
    & \textbullet
    & \textbullet
    & \textbullet 
    & \textbullet 
    & \textbullet
    \\
\midrule

$\lor$(\myroman{1}) 
    & $\varphi \lor \psi \rightarrow \varphi$
    & \textbullet
    & \textbullet
    & \textbullet 
    & \textbullet 
    & \textbullet
    \\
\midrule
    
$\lor$(\myroman{2}) 
    & $\varphi \lor \psi \rightarrow \psi$
    & \textbullet
    & \textbullet
    & \textbullet 
    & \textbullet 
    & \textbullet
    \\
\midrule
    
$\lor$(\myroman{3})
    & $(\chi \rightarrow \varphi) \rightarrow ((\chi \rightarrow \psi) \rightarrow (\chi \rightarrow \varphi \lor \psi))$
    & \textbullet
    & \textbullet
    & \textbullet 
    & \textbullet 
    & \textbullet
    \\

\midrule
        
Prelinearity
    & $(\varphi \rightarrow \psi)\lor(\psi \rightarrow \varphi)$
    & \textbullet
    & \textbullet
    & \textbullet 
    & \textbullet 
    & \textbullet
    \\
\midrule

EFQ
    & $\overline{0} \rightarrow \varphi$
    & \textbullet
    & \textbullet
    & \textbullet 
    & \textbullet 
    & \textbullet
    \\
\midrule

& $\varphi \land \psi \leftrightarrow \psi \land \varphi$
& \textopenbullet
& \textopenbullet
& \textopenbullet
& \textopenbullet
& \textopenbullet
\\
\midrule

& $(\varphi \land \psi) \land \chi \leftrightarrow \varphi \land (\psi \land \chi)$
& \textopenbullet
& \textopenbullet
& \textopenbullet
& \textopenbullet
& \textopenbullet
\\
\midrule

& $\varphi \lor \psi \leftrightarrow \psi \lor \varphi$
& \textopenbullet
& \textopenbullet
& \textopenbullet
& \textopenbullet
& \textopenbullet
\\
\midrule

& $(\varphi \lor \psi) \lor \chi \leftrightarrow \varphi \lor (\psi \lor \chi)$
& \textopenbullet
& \textopenbullet
& \textopenbullet
& \textopenbullet
& \textopenbullet
\\
\midrule

Contraction
    & $(\varphi \rightarrow(\varphi \rightarrow \psi)) \rightarrow(\varphi \rightarrow \psi)$
    & \textopenbullet
    &  
    & 
    & \textbullet
    & 
    \\
\midrule

Wajsberg
    & $\neg \neg \varphi \rightarrow \varphi$
    & \textopenbullet
    &
    & \textbullet
    &
    &
    \\
\midrule

& $\varphi \land \neg \varphi \rightarrow \overline{0}$
    & \textopenbullet
    & 
    &
    &
    & \textbullet
    \\
\midrule
    
 \myroman{11}    & $\neg \neg \chi \rightarrow ((( \varphi \land \chi) \rightarrow (\psi \land \chi)) \rightarrow (\phi \rightarrow \psi))$
    & \textopenbullet
    & 
    &
    &
    & \textbullet
    \\

\bottomrule  
\end{tabular}
\label{tab:fullaxioms}
\end{table}

%% file: writeup/7-D-baselineoperators.tex
\newpage
\clearpage

\section{Logical Operators of GQE, Query2Box, and BetaE} \label{ap:baselineconj}


\subsection{Conjunction}

GQE, Query2Box, BetaE represent queries as vectors, boxes (axis-aligned hyper-rectangles), and Beta distributions, respectively. 
Fig. \ref{fig:baselineconj} illustrates embedding-based conjunction operators of the three models, which takes embeddings of queries $q_1, q_2$ as input and produce an embedding for $q_1 \land q_2$.
Our analysis in Section \ref{sec:analysis} focuses on the nature of the geometry operations, and we thus omit deep architectures of the logical operators of these models from our analysis, such as the attention mechanism or deep sets \cite{DeepSets} for weighing sub-queries. We assume all the sub-queries should have identical weights.
The details about the original implementation are given below.

\subsubsection{GQE}
\begin{equation*}
    \mathcal{C}(\bsq_1, \bsq_2) = 
    \Psi (\text{NN}_k(\bsq_1, \bsq_2))
\end{equation*}
where $\bsq_1, \bsq_2 \in \mathbb{R}^d$, $\text{NN}_k$ is a $k$-layer feedforward neural network, $\Psi$ is a symmetric vector function, usually defined as the element-wise mean \cite{Query2Box}. 

\subsubsection{Query2Box}
In Query2Box, each query is represented by $d$ boxes, i.e., a vector $\bsq_{\text{CEN}} \in \mathbb{R}^d$ for box centers and a vector $\bsq_{\text{OFF} }\in \mathbb{R}^{+d}$ for box sizes.
The conjunction operator $\mathcal{C}_{\text{CEN}}, \mathcal{C}_{\text{OFF}}$ are defined as follows:
\begin{align*}
    \mathcal{C}_{\text{CEN}} (\bsq_1, \bsq_2) & = w_1 \bsq_{\text{CEN} 1} + w_2 \bsq_{\text{CEN} 1}; \quad w_i = \frac{\exp(\text{MLP}(\bsq_i))}{\sum_j \exp(\text{MLP}(\bsq_j))}, i=1,2\\
        \mathcal{C}_{\text{OFF}} (\bsq_1, \bsq_2) 
        & = \mathbf{w'} \circ \min (\bsq_{\text{OFF} 1}, \bsq_{\text{OFF} 2}) ; \quad \mathbf{w'} \ = \sigma(\text{DeepSets}(\bsq_{\text{OFF} 1}, \bsq_{\text{OFF} 2)})
\end{align*}
where $w_i$ is an input query weight using the attention mechanism, 
$\mathbf{w'}$ is learnable coefficients that shrink box sizes,
$\circ$ is the element-wise product, 
$\text{MLP}(\cdot): \mathbb{R}^{2d} \rightarrow \mathbb{R}^d$ is the Multi-Layer Perceptron,
$\sigma(\cdot)$ is the sigmoid function, 
DeepSets($\cdot$) 
is the permutation-invariant deep architecture \cite{DeepSets}, 
and both $\min(\cdot)$ and $\exp(\cdot)$ are applied in a dimension-wise manner.
When analyzing properties, we assume the deep architectures can work ideally, and $\mathcal{C}_{\text{OFF}}$ always finds the perfect intersection box size.

\subsubsection{BetaE}
In BetaE, each query is represented by $d$ Beta distributions. Each Beta distribution is shaped by an alpha parameter and a beta parameter. Thus a query is represented by a vector $\bsq_{\alpha} \in \mathbb{R}^{+d}$ for alpha parameters and a vector $\bsq_{\beta} \in \mathbb{R}^{+d}$ for beta parameters.
The conjunction operator $\mathcal{C}_{\alpha}, \mathcal{C}_{\beta}$ are defined as follows:
\begin{align*}
    \mathcal{C}_{\alpha}(
        \bsq_{\alpha 1}, \bsq_{\alpha 2}
    ) & = w_1 \bsq_{\alpha_1}  + w_2 \bsq_{\alpha_2} \\
    \mathcal{C}_{\beta}(
        \bsq_{\beta 1}, \bsq_{\beta 2}
    ) & = w_1 \bsq_{\alpha_1}  + w_2 \bsq_{\alpha_2}; \quad w_i = \frac{\exp(\text{MLP}(\bsq_i))}{\sum_j \exp(\text{MLP}(\bsq_j))}, i=1,2
\end{align*}
where $w_i$ is the input query weight computed by the attention mechanism:
    

\subsection{Negation}\label{sec:beta2fail}
Disjunction operators of GQE, Query2Box, and BetaE are described in Section \ref{sec:properties} and thus omitted here.
To the best of our knowledge, BetaE is the only existing model that could model negation.
The negation operator $\mathcal{N}_{\alpha}, \mathcal{N}_{\beta}$ of BetaE are defined as follows:
\begin{align*}
    \mathcal{N}_{\alpha}(\bsq)(i) = 1/\bsq_{\alpha}(i), \quad i=1,...,d\\
    \mathcal{N}_{\beta}(\bsq)(i) = 1/\bsq_{\beta}(i), \quad i=1,...,d
\end{align*}

\begin{figure}[h]
    \centering
    \includegraphics[width= 0.4 \linewidth]{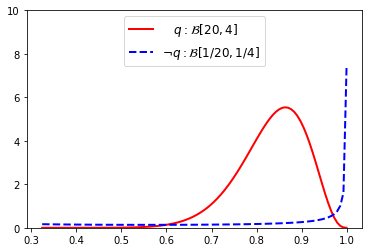}
    \caption{Illustration of an example where the negation operator of BetaE does not satisfy non-contradiction, i.e. $\phi(\neg q, e)$ is not monotonically decreasing with regard to $\phi(q, e)$.} 
    \label{fig:betaefail}
\end{figure}

Fig \ref{fig:betaefail} shows a case where the negation operator of BetaE does not satisfy non-contradiction, as $\phi(\neg q, e)$ is not monotonically decreasing with regard to $\phi(q, e)$.

%% file: writeup/7-E-dataset.tex
\newpage
\clearpage
\section{Dataset statistics and query structures} \label{ap:datasets}
We use the benchmark datasets provided by BetaE \cite{BetaE}, which is publicly available at \url{https://github.com/snap-stanford/KGReasoning}.
The datasets contain 14 types of logical queries on 
FB15k-237 \cite{FB15k237} and NELL995 \cite{DeepPath} respectively. 
The queries are generated based on the official training/validation/testing edge splits of those KGs.
The 14 types of query structures in the datasets are shown in Fig. \ref{fig:query_structure}.
We list the number of training/validation/test queries in Table \ref{table:querynums}.
The KG statistics are summarized in Table \ref{table:kgstats}.

\input{tables/kgstats}

\input{tables/querynums}

%% file: tables/kgstats.tex
\begin{table*}[h]
\centering
\caption{Knowledge graph dataset statistics as well as training, validation, and test edge splits.}
\label{table:kgstats}
\begin{tabular}{ccccccc}
\toprule
   Dataset &  Entities &  Relations &  Training Edges &  Validation Edges &  Test Edges &  Total Edges \\
\midrule
 FB15k-237 &     14505 &        237 &          272115 &             17526 &       20438 &       310079 \\
      NELL &     63361 &        200 &          114213 &            143234 &       14267 &       142804 \\
\bottomrule
\end{tabular}
\end{table*}

%% file: tables/querynums.tex
\begin{table*}[h]
\centering
\caption{Number of training, validation, and test queries for different query structures. For columns that list multiple query structures, the number in the table represents the number of each query structure.}
\label{table:querynums}
\begin{tabular}{ccccccc}
\toprule
    &      \multicolumn{2}{c}{Training}  &  \multicolumn{2}{c}{Validation}&   \multicolumn{2}{c}{Test} \\
\cmidrule(lr){1-1} \cmidrule(lr){2-3}  \cmidrule(lr){4-5} \cmidrule(lr){6-7}
   Dataset &  1p/2p/3p/2i/3i &  2in/3in/inp/pin/pni &         1p &       others &     1p &  others \\
\midrule
 FB15k-237 &          149,689 &                149,68 &      20,101 &         5,000 &  2,812 &    5,000 \\
   NELL995 &          107,982 &                10,798 &      16,927 &         4,000 &  17,034 &    4,000 \\
\bottomrule
\end{tabular}
\end{table*}

%% file: writeup/7-F-experiment.tex
\newpage
\clearpage

\section{Experimental details} \label{ap:experiments}

\subsubsection{Implementation}
For GQE \cite{GQE}, Query2Box \cite{Query2Box}, and BetaE \cite{BetaE}, we use the implementation from \url{https://github.com/snap-stanford/KGReasoning}.
For CQD, we use the implementation at \url{https://github.com/uclnlp/cqd}.
The source code of our model is included in a code appendix. 
All source code of our model will be made publicly available upon publication of the paper with a license that allows free usage for research purposes.

\subsubsection{Model configurations and hyperparameters}
We use AdamW \cite{AdamW} as the optimizer. Training terminates with early stopping based on the average MRR on the validation set with a patience of 15k steps. We run each method up to 450k steps.
We repeat each experiment three times and report the average results.

As in \cite{Query2Box, BetaE}, for fair comparison, we use the same embedding dimensionality $d$ and the number of negative samples $k$ for all the methods.
With reference to \cite{BetaE}, we set the embdding dimensionality to $d=800$ and use $k=128$ negative samples per positive sample.
We finetune other hyperparameters and the choice of the subspace mapping function $\mathbf{g}: \mathbb{R}^d \rightarrow [0,1]^d$ by grid search based on the average MRR on the validation set.
We search hyperparameters in the following range: 
learning rate from $\{0.001, 0.0005, 0.0001\}$,
number of relation bases from $\{30, 50, 100, 150\}$,
batch size $b$ from \{128, 512, 1000\}.
$\mathbf{g}$ is chosen from from \{Logistic function, Bounded rectifier\}.

The best hyperparameter combination on FB15k-237 is learning rate $0.001$, number of relation bases $150$, batch size $512$, $\mathbf{g}$ as a logistic function.
The best combination on NELL995 is learning rate $0.0005$, number of relation bases $30$, batch size $1000$, $\mathbf{g}$ as a bounded rectifier.
For baselines GQE , Q2B, and BetaE, we use the best combinations reported by \cite{BetaE}. For CQD, we use the ones reported in \cite{CQD}.
We follow the setting in the official code repository for any hyperparameter unspecified in the paper.

\subsubsection{Hardware and software Specifications} 
Each single experiment is run on CPU  $\text{Intel}^{\circledR}$ $\text{Xeon}^{\circledR}$ E5-2650 v4 12-core and a single  $\text{NVIDIA}^{\circledR}$ GP102 TITAN Xp (12GB) GPU. RAM size is 256GB.
The operating system is Ubuntu 18.04.01. Our framework is implemented wtih Python 3.9 and Pytorch 1.9.


%% file: writeup/7-G-fuzzylogic.tex
\newpage
\clearpage

\section{\tnorm~based fuzzy logic systems} \label{ap:tnorm}


Functions that qualify as fuzzy conjunction and fuzzy disjunction are usually referred to in literature as \tnorms~ (triangular norms) and \tconorms~ (triangular conorms) respectively in literature \cite{KlirFuzzyBook}.


A \tnorm~\cite{tnorm} is a function $t: [0, 1] \times [0, 1] \rightarrow [0, 1]$ which is commutative and associative and satisfies the boundary condition $t(a, 1) = a$ as well as monotonicity, i.e., $t(a, b) \leq t(c, d)$ if $a \leq c$ and $b \leq d$.
Prominent examples of \tnorms~include minimum, product, and \luka~\tnorm. 

\tconorm~is the logical dual form of the \tnorm. Following De Morgan's law, given the canonical negator $c(x)=1-x$, the \tconorm~ $s$ of a \tnorm~$t$ is defined as
$s(a,b) =  1 - t(1-a, 1-b)$.
A \tconorm~is commutative and associative, and satisfies the boundary condition $s(a, 0) = a$ as well as monotonicity: $s(a, b) \leq s(c, d)$ if $a \leq c$ and $b \leq d$. Interested readers are referred to \cite{KlirFuzzyBook} for proofs.

\begin{figure*}[h]
    \centering
    \includegraphics[width=\linewidth]{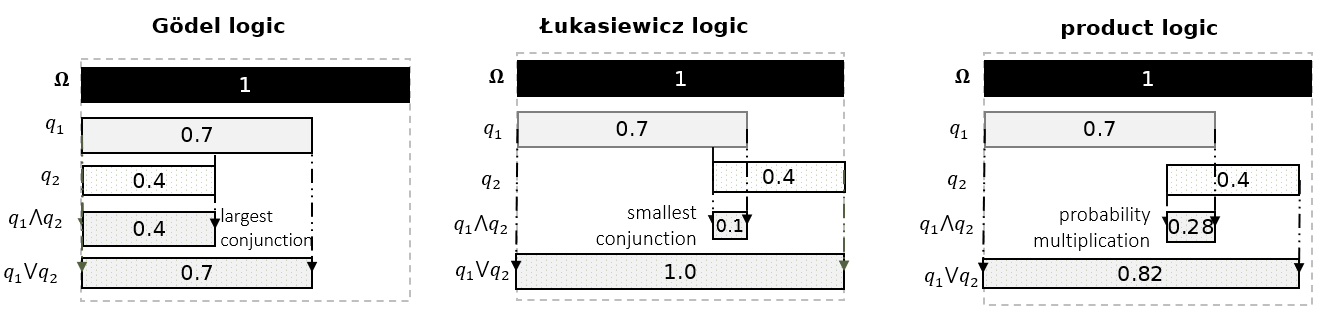}
    \caption{Illustration of fuzzy conjunction and disjunction, which is equivalent to fuzzy set intersection and union.
    } 
    \label{fig:fuzzylogic}
\end{figure*}

The formulas of \tconorms~that correspond to the minimum (\godel), product, and \luka~\tnorms~are given in Table \ref{tab:tnorms}. The illustration is given in Fig \ref{fig:fuzzylogic}.

\input{tables/t-norms}

%% file: tables/t-norms.tex
\begin{table*}[h]
\centering
\caption{Prominent examples of \tnorms~and the corresponding \tnorms~derived by De Morgan's law and the canonical negator $c(x)=1-x$. We list the special properties of the formulas in addition to the basic properties (i.e., commutativity, associativity, monotonicity, and boundary condition) of t-norm and t-conorm.}
\begin{tabular}{llll}
\toprule
\multicolumn{2}{c}{\tnorm~($\land$)}  & \tconorm~($\lor$) & Special Properties \\
\cmidrule(lr){1-2} \cmidrule(lr){3-3} \cmidrule(lr){4-4} 

minimum (\godel) & $t(a,b)=\min(a,b)$
&  $s(a,b)=\max(a,b)$
& idempotent
\\

product & $t(a,b)=ab$
& $s(a,b)=a+b-ab$ 
& strict monotonicity
\\

\luka & $t(a,b)=\max(a+b-1, 0)$
& $s(a,b)=\min(a+b, 1)$ 
& nilpotent
\\

\bottomrule
\end{tabular}

\label{tab:tnorms}
\end{table*}

%% file: writeup/7-H-time.tex
\newpage
\clearpage
\section{Time Comparison with CQD} \label{sec:time}

We compare with CQD with regard to the time for answering a query. 
The experiment is conducted on a  $\text{NVIDIA}^{\circledR}$ GP102 TITAN Xp (12GB) GPU. The full hardware and software specifications are given in Appendix \ref{ap:experiments}.
For CQD, we use its official implementation \footnote{https://github.com/uclnlp/cqd} and experiment setting \cite{CQD}. The beam search candidate number for CQD is set as $64$, i.e., CQD finds top $64$ entity candidates for each sub-query and uses it as seeds for search in the next round.
For \modelname, we retrieve top $64$ entity candidates for each query as well. 
We use FAISS \cite{FAISS} to speed up dense similarity search, where\emph{exact measurement matching} is adopted instead of \emph{approximate measurement matching}.
FAISS cannot be applied to CQD, because (i) CQD is not a logical query embedding framework that retrieves entity answers by dense similarity search, and (ii) scoring an entity for a query involves computation in the complex number domain.

\begin{figure}[h]
    \centering
    \includegraphics[width=0.6 \linewidth]{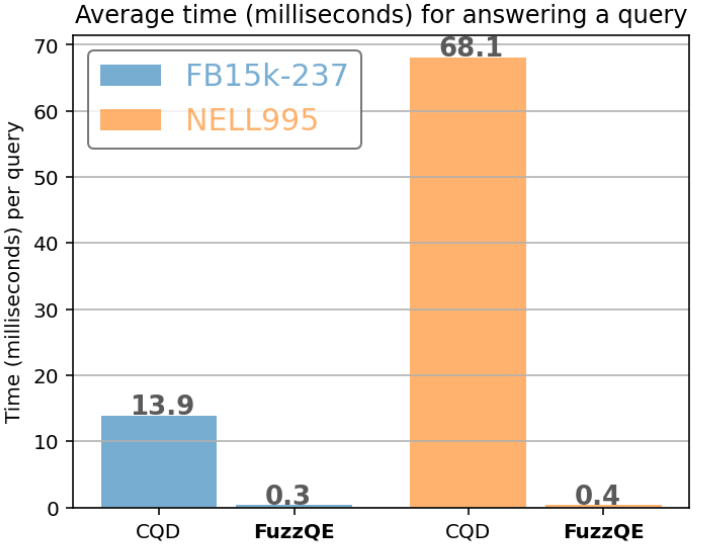}
    \caption{Average time (milliseconds) for answering an FOL query on a single  $\text{NVIDIA}^{\circledR}$ GP102 TITAN Xp (12GB) GPU. FB15k-237 contains 14,505 entities. NELL995 contains 63,361 entities, roughly 4 times the number of FB15k-237.} 
    \label{fig:inferencetime}
\end{figure}

\begin{figure}[h]
    \centering
    \includegraphics[width=\linewidth]{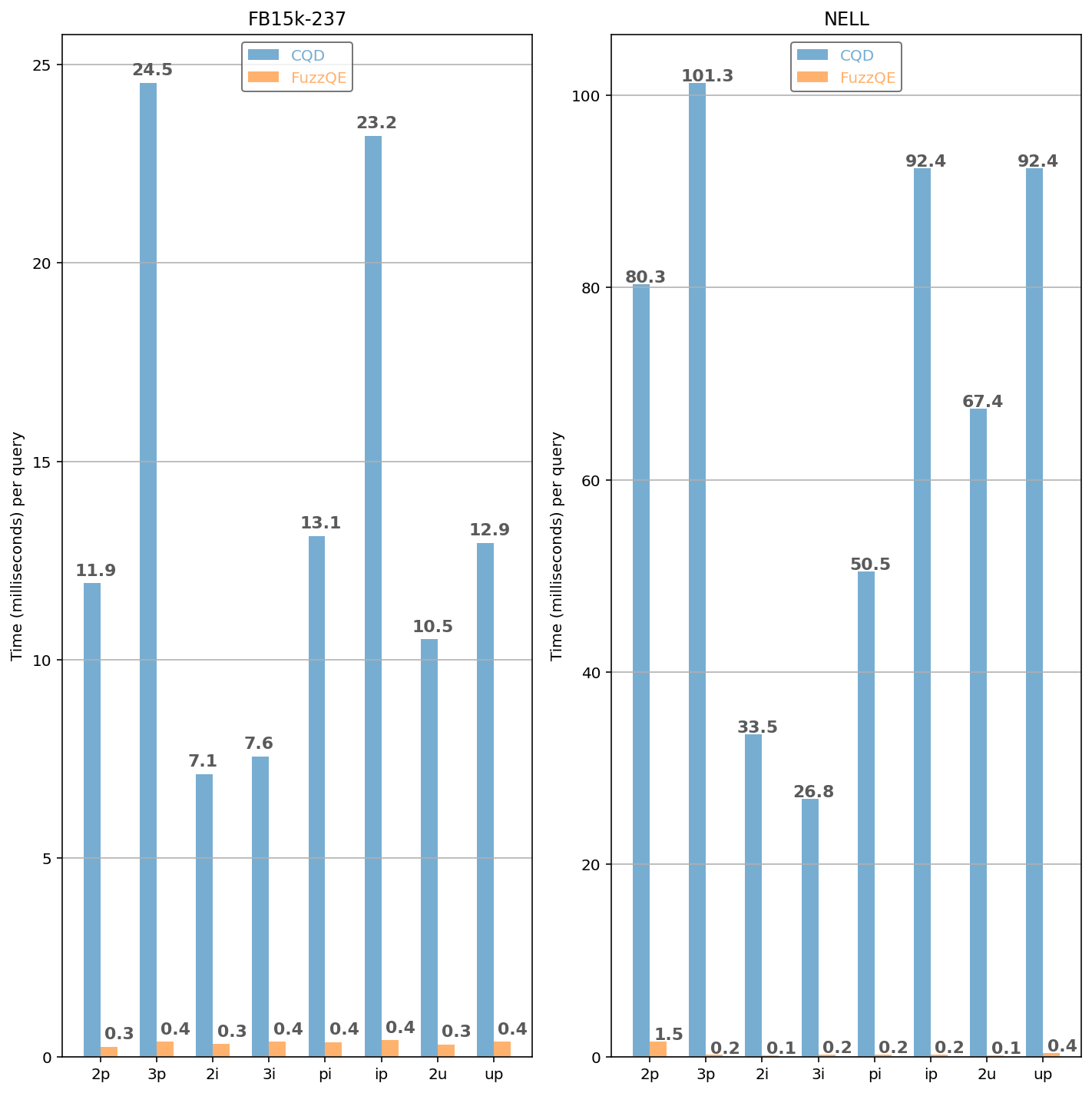}
    \caption{Average time (milliseconds) required by CQD and FuzzQE for answering each query type in FB15k-237 and NELL995.} 
    \label{fig:timebytype}
\end{figure}

Fig \ref{fig:inferencetime} shows the average time of CQD and FuzzQE for answering a complex FOL query. 
Fig \ref{fig:timebytype} shows the time required by CQD and FuzzQE for answering each query type, aggregated over FB15k-237 and NELL995. The structure of different query types are shown in Fig. \ref{fig:query_structure} in Appendix \ref{ap:datasets}.
Consistent with the observation in \cite{CQD}, the main computation bottleneck of CQD are multi-hop queries (e.g., $3p$ queries), since the model is required to invoke the link prediction model for each node in the dependency graph to obtain the top-$k$ candidates for the next step.
We also note that as the number of entities increases, the time required by CQD to answer a query significantly grows.
In contrast, the inference time of \modelname is almost independent of the number of entities and the complexity of the query.

